\documentclass{article}
\usepackage{spconf,amsmath,graphicx}

\usepackage[utf8]{inputenc}
\usepackage[T1]{fontenc}
\usepackage{hyperref}
\usepackage{url}
\usepackage{booktabs}
\usepackage{amsfonts}
\usepackage{nicefrac} 
\usepackage{microtype}

\usepackage{adjustbox} 

\usepackage{booktabs}
\usepackage{wrapfig}
\usepackage{array}
\usepackage{tablefootnote}
\usepackage{color}
\usepackage{cite}

\usepackage{xcolor}         
\usepackage{placeins}
\usepackage{caption}
\usepackage{subcaption}

\usepackage{xcolor,pifont}
\newcommand*\colourcheck[1]{%
  \expandafter\newcommand\csname #1check\endcsname{\textcolor{#1}{\ding{52}}}%
}
\colourcheck{blue}
\colourcheck{green}
\colourcheck{red}
\newcommand*\colourcross[1]{%
  \expandafter\newcommand\csname #1cross\endcsname{\textcolor{#1}{\ding{56}}}%
}
\colourcross{red}

\usepackage{wrapfig}

\usepackage{multirow,mathtools} \usepackage{algorithm,algpseudocode}

\usepackage{pifont}
\usepackage{color, colortbl}

\usepackage{blindtext}
\usepackage{lipsum}
\usepackage{threeparttable}

\usepackage{amsmath,amsfonts,bm}

\def\eqref#1{(\ref{#1})}

\def\1{\bm{1}}

\DeclareMathAlphabet{\mathsfit}{\encodingdefault}{\sfdefault}{m}{sl}
\SetMathAlphabet{\mathsfit}{bold}{\encodingdefault}{\sfdefault}{bx}{n}

\DeclareMathOperator*{\argmax}{arg\,max}

\DeclareMathOperator*{\minimize}{\text{minimize}}
\DeclareMathOperator*{\maximize}{\text{maximize}}

\newcommand{\btheta}{\boldsymbol{\theta}}

\usepackage{algorithm,algpseudocode}
\algnewcommand{\algorithmicforeach}{\textbf{for each}}
\algdef{SE}[FOR]{ForEach}{EndForEach}[1]
  {\algorithmicforeach\ #1\ \algorithmicdo}
  {\algorithmicend\ \algorithmicforeach}

\usepackage{pifont}

\usepackage{color, colortbl}
\definecolor{Gray}{gray}{0.93}
\definecolor{Orange}{rgb}{1,0.5,0}
\definecolor{DGray}{gray}{0.83}
\definecolor{LightCyan}{rgb}{0.88,1,1}

\usepackage[most]{tcolorbox}

\usepackage{amsmath}
\usepackage{amssymb}
\usepackage{amsfonts}
\usepackage{graphicx}
\usepackage{wrapfig}
\usepackage{times}
\usepackage{multirow}
\usepackage{url}
\usepackage{color}
\usepackage{xcolor}

\usepackage{enumitem}
\usepackage{xspace}
\usepackage{xfrac}
\usepackage{tcolorbox}
\usepackage{comment}
\usepackage{tikz}
\usepackage{etoolbox}

\definecolor{SL_color}{rgb}{0.858, 0.188, 0.478}

\usepackage{pifont}
\usepackage{adjustbox}
\usepackage{threeparttable,booktabs}
\usepackage{multirow}

\DeclareMathAlphabet\mathbfcal{OMS}{cmsy}{b}{n}

\DeclareMathOperator*{\ST}{\text{subject to}}

\usepackage{amsmath}%
\usepackage{MnSymbol}%
\usepackage{wasysym}%

\usepackage{tcolorbox}

\usepackage{mathtools}
\DeclarePairedDelimiterX{\inp}[2]{\langle}{\rangle}{#1, #2}

\usepackage{color, colortbl}
\definecolor{Gray}{gray}{0.93}
\definecolor{Orange}{rgb}{1,0.5,0}
\definecolor{DGray}{gray}{0.83}
\definecolor{LightCyan}{rgb}{0.88,1,1}

\usepackage{{tcolorbox}}

\title{
Visual Prompting for Adversarial Robustness
}
\name{Aochuan Chen$^{\star, 1}$ \qquad Peter Lorenz$^{\star, 2}$\qquad Yuguang Yao$^{1}$\qquad Pin-Yu Chen$^{3}$\qquad Sijia Liu$^{1}$}
\address{
$^{1}$Michigan State University, USA\\
$^{2}$Fraunhofer ITWM and Fraunhofer Center of Machine Learning, Germany\\
$^{3}$IBM Research, USA
}

\begin{document}

\maketitle
\let\thefootnote\relax\footnotetext{$^{\star}$ Equal contribution.}

\begin{abstract}
In this work, we leverage visual prompting (VP) to improve adversarial robustness of a fixed, pre-trained  model at test time. Compared to conventional adversarial defenses, VP allows us to design universal  (\textit{i.e.}, data-agnostic) input prompting templates, which have   plug-and-play capabilities at test time to achieve   desired model  performance without introducing much  computation overhead.
Although VP has been successfully applied   to improving model generalization, it remains  elusive whether and how it can be used to defend against adversarial attacks.
We investigate this problem and show that the vanilla VP approach is \textit{not} effective in adversarial defense since a universal input prompt lacks the capacity for robust learning against sample-specific adversarial perturbations.   
To circumvent it,  we propose a new VP method, termed \underline{C}lass-wise \underline{A}dversarial \underline{V}isual \underline{P}rompting (C-AVP), to generate class-wise visual prompts so as to not only leverage  the strengths of ensemble prompts but also optimize their interrelations to improve model robustness.  
Our experiments show that C-AVP outperforms the conventional VP method, with 2.1$\times$ standard accuracy gain  and 2$\times$ robust accuracy gain. 
Compared to classical test-time defenses,  C-AVP also yields a 42$\times$ inference time speedup. Code is available at \textit{https://github.com/Phoveran/vp-for-adversarial-robustness}. 
\end{abstract}


\vspace*{-4mm}
\section{Introduction}
\vspace*{-2mm}
Machine learning (ML) models can easily be manipulated (by an adversary) to output drastically different classifications.
Thereby, model robustification against adversarial attacks is now a major focus of research. Yet, a large volume of existing works focused on training recipes and/or model architectures to gain robustness. Adversarial training (AT) \cite{madry2017towards}, one of the most  effective defense, adopted min-max optimization to minimize  the worst-case training loss induced by adversarial attacks. Extended from AT, various defense methods were proposed, ranging from supervised learning to semi-supervised learning, and further to unsupervised learning \cite{zhang2019theoretically,shafahi2019adversarial,zhang2019you,carmon2019unlabeled,wong2017provable, raghunathan2018certified,xie2019feature,chen2020adversarial,fan2021does, jia2022clawsat}. 

Although the design for robust training   has made tremendous success in   improving model robustness \cite{athalye2018obfuscated,croce2020reliable},  it typically takes an intensive computation cost with poor defense scalability to  a fixed, pre-trained ML model.
Towards circumventing this difficulty, the problem of test-time defense arises; see the seminal work in Croce \textit{et. al.} \cite{croce2022evaluating}. Test-time defense alters either a test-time input example or a small portion of the pre-trained model.  Examples include  input (anti-adversarial) purification \cite{yoon2021adversarial,mao2021adversarial,alfarra2022combating} and model refinement by augmenting the pre-trained model with auxiliary components \cite{salman2020denoised,gong2022reverse,kang2021stable}. However, these  defense techniques inevitably raise the inference
time and hamper the test-time efficiency \cite{croce2022evaluating}. 
Inspired by that, our work will advance the test-time defense technology by leveraging the idea of \textit{visual prompting} (\textbf{VP}) 
\cite{bahng2022visual},  also known as  model reprogramming \cite{chen2022model,elsayed2018adversarial,tsai2020transfer,zhang2022fairness}.

Generally speaking, VP \cite{bahng2022visual} creates a \textit{universal} (\textit{i.e.}, \textit{data-agnostic}) input prompting template  (in terms of  input perturbations) in order to improve the generalization ability of a pre-trained   model when incorporating such a visual prompt into test-time examples. 
It enjoys the same idea as {model reprogramming}  \cite{chen2022model,elsayed2018adversarial,tsai2020transfer,zhang2022fairness} or {unadversarial example} \cite{salman2021unadversarial}, which optimizes a universal perturbation pattern to maneuver (\textit{i.e.}, reprogram) the functionality of a pre-trained model towards the desired criterion, \textit{e.g.},  cross-domain transfer learning \cite{tsai2020transfer},
out-of-distribution generalization \cite{salman2021unadversarial},
and fairness \cite{zhang2022fairness}. 
However, it remains elusive  whether or not VP could be designed as an effective solution to adversarial defense. We will investigate this problem, which we call \textit{adversarial visual prompting} (\textbf{AVP}) in this work. Compared to conventional test-time defense methods, AVP significantly reduces the inference time overhead since visual prompts can be designed offline over training data and have the plug-and-play capability  applied to any testing data. We summarize our \textbf{contributions} as below. \\
\ding{182} We formulate and investigate the problem of AVP for the first time and empirically show the conventional data-agnostic VP design is incapable of gaining adversarial robustness. \\
\ding{183} We propose a new VP method, termed class-wise AVP (\textbf{C-AVP}), which produces multiple, class-wise visual prompts with explicit optimization on their couplings to gain better adversarial robustness. \\
\ding{184} We provide insightful experiments to demonstrate the pros and cons of VP in adversarial defense. 

\section{Related work}
\noindent \textbf{Visual prompting.}
Originated from the idea of in-context learning or prompting in natural language processing (NLP)  \cite{brown2020language,li2021prefix,radford2021learning,zhang2023text}, VP was first proposed in Bahng \textit{et. al.}\cite{bahng2022visual} for vision models. 
Before formalizing VP in Bahng \textit{et. al.}\cite{bahng2022visual}, the underlying prompting technique has also  been devised in computer vision (CV) with  different naming. For example, VP is closely related to \textit{adversarial reprogramming}  or \textit{model reprogramming} \cite{elsayed2018adversarial,chen2022model,tsai2020transfer,neekhara2022cross,yang2021voice2series,zheng2021adversarial}, which focused on altering the functionality of a fixed, pre-trained model {across domains} by augmenting test-time examples with an additional (universal) input perturbation pattern. \textit{Unadversarial learning} also enjoys the  similar idea to VP.  In \cite{salman2021unadversarial}, unadversarial examples that perturb original ones using  `prompting' templates were introduced to improve  out-of-distribution generalization.
Yet, the problem of VP for adversarial defense is under-explored. 

\noindent \textbf{Adversarial defense.}
The lack of adversarial robustness is
a weakness of  ML models. Adversarial defense, such as adversarial detection \cite{grosse2017statistical, yang2019ml, metzen2017detecting, meng2017magnet, wojcik2020adversarial, gong2022reverse}   and robust training \cite{wong2017provable, zhang2019theoretically, salman2020denoised, chen2020adversarial, boopathy2020proper, fan2021does}, is a current research focus. In particular, adversarial training (AT) \cite{madry2017towards} is the most widely-used defense strategy and has inspired many recent advances in adversarial defense \cite{athalye2018obfuscated, ye2019adversarial, croce2020reliable, mohapatra2020rethinking, kang2021stable, wang2021on}.  However, these AT-type defenses (with the goal of robustness-enhanced model  training) are   computationally intensive due to   min-max optimization over model parameters.  
To reduce the computation overhead of robust training, the problem of test-time defense arises \cite{croce2022evaluating}, which aims to robustify a   given model  via lightweight unadversarial input perturbations (\textit{a.k.a} input purification)  \cite{shi2021online, yoon2021adversarial}   or minor modifications to the fixed model \cite{chen2021towards, zhang2022robustify}.  In different kinds of test-time defenses, the most relevant work to ours is  anti-adversarial perturbation \cite{alfarra2022combating}. 

\vspace*{-4mm}
\section{Problem Statement}
\vspace*{-3mm}
\noindent \textbf{Visual prompting.}
We describe the problem setup of VP following Bahng \textit{et. al.}\cite{bahng2022visual,elsayed2018adversarial,tsai2020transfer,zhang2022fairness}.
Specifically, let $\mathcal D_\mathrm{tr}$ 
denote a  training set for supervised learning, where $(\mathbf x, y) \in \mathcal D_\mathrm{tr}$ signifies a training  sample with feature $\mathbf x$ and label $y$. And let $\boldsymbol \delta$ be a visual prompt to be designed. The prompted input is then given by $\mathbf x + \boldsymbol \delta$ with respect to (w.r.t.) $\mathbf x$. 
Different from the problem of adversarial attack generation that optimizes  $\boldsymbol \delta$ for erroneous prediction, VP drives $\boldsymbol \delta$ to minimize the performance loss $\ell$ of a pre-trained model $\boldsymbol \theta$. This leads to

{\vspace*{-4.5mm}
    \small 
    \begin{align}
        \begin{array}{ll}
    \displaystyle \minimize_{\boldsymbol \delta}         &  \mathbb E_{(\mathbf x, y) \in \mathcal D_\mathrm{tr}} [ \ell ( \mathbf{x} + \boldsymbol \delta; y, \boldsymbol \theta ) ]\\
          \ST    & \boldsymbol \delta \in \mathcal C,
        \end{array}
        \label{eq: visual_prompt}
    \end{align}
}%
where $\ell$ denotes prediction error given the training data $(\mathbf x, y)$ and   base model $\boldsymbol \theta$, and $\mathcal C$ is a perturbation constraint. 
Following Bahng \textit{et. al.}\cite{bahng2022visual,elsayed2018adversarial,tsai2020transfer},  $\mathcal C$  restricts  $\boldsymbol \delta$ to let $ \mathbf x + \boldsymbol \delta \in  [0, 1]$ for any $\mathbf x$. 
Projected gradient descent (PGD) \cite{madry2017towards,salman2021unadversarial} can then be applied to
 solving problem \eqref{eq: visual_prompt}. 
In the evaluation, $\boldsymbol \delta$ is integrated into  test data to improve the prediction ability of $\boldsymbol \theta$.

\noindent \textbf{Adversarial visual prompting.}
Inspired by the usefulness of VP to improve model generalization \cite{tsai2020transfer,bahng2022visual}, we ask: 
\begin{center}
	\vspace*{-5.5mm}
	\setlength\fboxrule{0.1pt}
	\noindent\fcolorbox{black}[rgb]{0.85,0.9,0.95}{\begin{minipage}{0.98\columnwidth}
				\vspace{-0.00cm}
	\textbf{(AVP problem)}	
	Can VP \eqref{eq: visual_prompt} be extended to robustify $\boldsymbol \theta$ against adversarial attacks?
				\vspace{-0.00cm}
	\end{minipage}}
	\vspace*{-1.5mm}
\end{center}
At the first glance, the AVP problem seems trivial if we specify the performance loss $\ell$ as the adversarial training loss \cite{madry2017towards,zhang2019theoretically}:

{
\vspace*{-4.5mm}
\small 
\begin{align}
    \ell_{\mathrm{adv}}(\mathbf x + \boldsymbol \delta; y , \boldsymbol \theta) =  \maximize_{\mathbf x^\prime: \| \mathbf x^\prime - \mathbf x \|_\infty \leq \epsilon }   \ell(\mathbf x^\prime + \boldsymbol \delta; y , \boldsymbol \theta),
    \label{eq: adv_loss}
\end{align}}%
where $\mathbf x^\prime$ denotes the adversarial input that lies in the $\ell_\infty$-norm ball centered at  $\mathbf x$ with radius $\epsilon > 0$.

Recall from \eqref{eq: visual_prompt} that the conventional VP requests $\boldsymbol \delta$ to be  universal across training data. Thus, we term 
\textit{universal AVP} (\textbf{U-AVP}) 
the following problem by integrating \eqref{eq: visual_prompt} with \eqref{eq: adv_loss}:

\vspace*{-4.5mm}
{\small \begin{align}
    \begin{array}{ll}
    \displaystyle \minimize_{\boldsymbol \delta: \, \boldsymbol \delta \in \mathcal C}         &  
    \lambda \mathbb E_{(\mathbf x, y) \in \mathcal D_\mathrm{tr}} [ \ell ( \mathbf{x} + \boldsymbol \delta; y, \boldsymbol \theta ) ] + \\  &
    \mathbb E_{(\mathbf x, y) \in \mathcal D_\mathrm{tr}} [ \ell_{\mathrm{adv}}(\mathbf x + \boldsymbol \delta; y , \boldsymbol \theta)  ]
    \end{array}
    \label{eq: U_AVP}
    \tag{U-AVP}
\end{align}}%
where $\lambda > 0$ is a regularization parameter to strike a balance between generalization and adversarial robustness \cite{zhang2019theoretically}.

\begin{wrapfigure}{i}{.5\columnwidth}
    \centering
    \vspace*{-6mm}
    \includegraphics[width=.48\columnwidth]{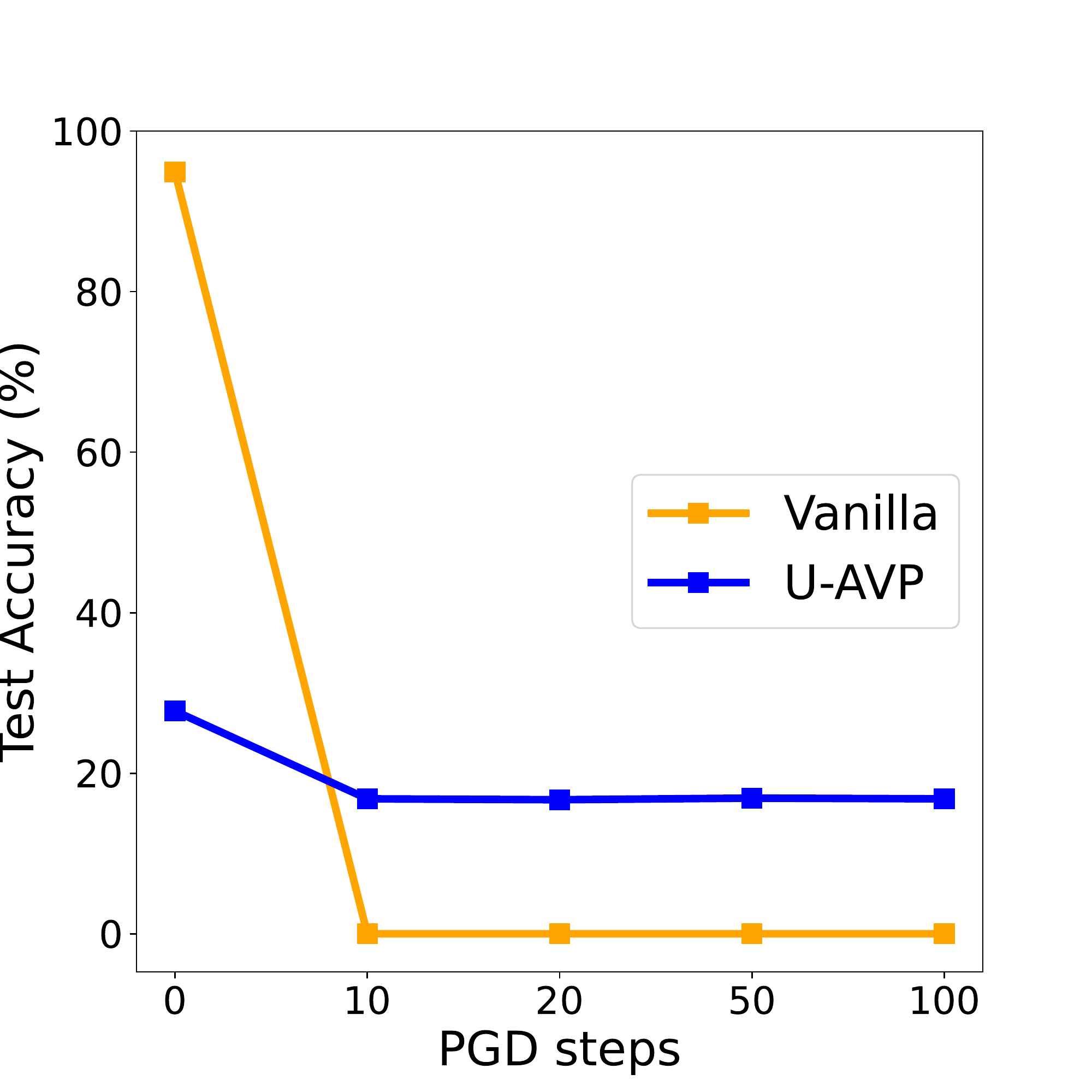}
    \caption{\footnotesize{Example of designing U-AVP for adversarial defense on (CIFAR-10, ResNet18), measured by robust accuracy against PGD attacks \cite{madry2017towards} of different steps. The robust accuracy of $0$ steps is the standard accuracy. 
    }}
    \vspace*{-4mm}
    \label{fig: warm-up: U-AVP}
\end{wrapfigure}
The problem \eqref{eq: U_AVP} can be effectively solved using  a standard min-max optimization method, which involves two alternating optimization routines:  inner maximization and outer minimization. The former generates adversarial examples as AT, and the latter produces the visual prompt $\boldsymbol \delta$ like \eqref{eq: visual_prompt}. 
At test time, the effectiveness of $\boldsymbol \delta$ is measured from two aspects: (1) standard accuracy, \textit{i.e.}, the accuracy of $\boldsymbol \delta$-integrated benign examples, and (2) robust accuracy, \textit{i.e.}, the accuracy of $\boldsymbol \delta$-integrated adversarial examples (against the victim model $\boldsymbol \theta$).
Despite the succinctness of  \eqref{eq: U_AVP},   Fig.\,\ref{fig: warm-up: U-AVP} 
shows its \textit{ineffectiveness}  to defend against adversarial attacks. Compared to the vanilla VP \eqref{eq: visual_prompt}, it also suffers a significant standard accuracy drop (over $50\%$ in Fig.\,\ref{fig: warm-up: U-AVP} corresponding to $0$ PGD attack steps) and robust accuracy is only enhanced by a small margin (around $18\%$ against   PGD attacks). The negative results in Fig.\,\ref{fig: warm-up: U-AVP} are not   quite surprising since a data-agnostic input prompt $\boldsymbol \delta$ has limited learning capacity to enable adversarial defense. Thus, it is non-trivial to tackle the problem of AVP. 

\vspace*{-4mm}
\section{Class-wise Adversarial Visual Prompt}
\vspace*{-3mm}
\noindent \textbf{No free lunch for class-wise visual prompts.}
A direct extension of \eqref{eq: U_AVP} is to introduce multiple adversarial visual prompts, each of which corresponds to one class in the training set $\mathcal D_\mathrm{tr}$. 
If we split $\mathcal D_\mathrm{tr}$ into class-wise training sets $\{ \mathcal{D}_\mathrm{tr}^{(i)}\}_{i=1}^N$ (for $N$ classes) and introduce class-wise visual prompts $\{ \boldsymbol \delta^{(i)} \}$, then  the direct C-AVP extension from \eqref{eq: U_AVP} becomes

{\vspace*{-4.5mm} 
\small
\begin{align}
     \begin{array}{ll}
    \displaystyle \minimize_{ \{ \boldsymbol \delta^{(i)} \in \mathcal C\}_{i\in [N]}}          & \displaystyle \frac{1}{N}\sum_{i=1}^N \left \{
         \lambda \mathbb E_{(\mathbf x, y) \in \mathcal D_\mathrm{tr}^{(i)}} [ \ell ( \mathbf{x} + \boldsymbol \delta^{(i)}; y, \boldsymbol \theta ) ] + \right. \\ & \left. 
        \mathbb E_{(\mathbf x, y) \in \mathcal D_\mathrm{tr}^{(i)}} [ \ell_{\mathrm{adv}}(\mathbf x + \boldsymbol \delta^{(i)}; y , \boldsymbol \theta)  ] \right \}
    \end{array}
    \label{eq: C_AVP_vanilla}
    \tag{C-AVP-v0}
\end{align}}%
where $[N]$ denotes the set of class labels $\{ 1,2,\ldots, N \}$.
It is worth noting that   \ref{eq: C_AVP_vanilla} is \textit{decomposed} over class labels. 
Although the  class-wise separability facilitates numerical optimization, it introduces challenges \textbf{(C1)}-\textbf{(C2)} when applying class-wise visual prompts for adversarial defense.

$\bullet$ \textbf{(C1)} \textit{Test-time prompt selection}: After acquiring the  visual prompts $\{ \boldsymbol \delta^{(i)}\}$  from \eqref{eq: C_AVP_vanilla}, it remains unclear  how     a class-wise   prompt should be selected  for application to a test-time example $\mathbf x_\mathrm{test}$. 
An intuitive way is to use the   inference pipeline of   $\boldsymbol \theta$ by aligning its top-$1$ prediction with  the prompt selection. That is, the selected prompt $\boldsymbol \delta$ and the predicted class $i^*$  are determined by

{
\vspace*{-4.5mm}
\small \begin{align}
 \boldsymbol \delta = \boldsymbol \delta^{*} , ~    i^{*} = \argmax_{i \in [N]} f_i (\mathbf x_\mathrm{test} + \boldsymbol \delta^{(i)}; \boldsymbol \theta),
 \label{eq: rule_prompt_sel}
\end{align}}%
where $f_i(\mathbf x; \btheta)$ denotes the $i$th-class prediction confidence.
However, the seemingly correct rule \eqref{eq: rule_prompt_sel} leads to a large prompt selection error (thus poor prediction accuracy) due to \textbf{(C2)}.

$\bullet$ \textbf{(C2)} \textit{Backdoor effect of class mis-matched prompts}: 
Given $\boldsymbol \delta^{(i)}$ from \eqref{eq: C_AVP_vanilla}, if the test-time example $\mathbf x_\mathrm{test}$ is drawn from class $i$,   
the visual prompt $\boldsymbol \delta^{(i)}$ then helps    prediction. However, if $\mathbf x_\mathrm{test}$ is \textit{not} originated from class $i$, then $\boldsymbol \delta^{(i)}$  could serve as a backdoor attack trigger \cite{gu2017badnets} with the targeted backdoor label  $i$ for   the `prompted input' $\mathbf x_\mathrm{test} + \boldsymbol \delta^{(i)}$. 
Since the backdoor attack is also input-agnostic, the class-discriminative ability of $\mathbf x_\mathrm{test} + \boldsymbol \delta^{(i)}$ enabled  by $\boldsymbol \delta^{(i)}$ could result in  incorrect prediction  towards the target class $i$ for  $\mathbf x_\mathrm{test} $.

\begin{figure}[htp]
    \centering
    \vspace*{-2mm}
    \includegraphics[width=.46\textwidth]{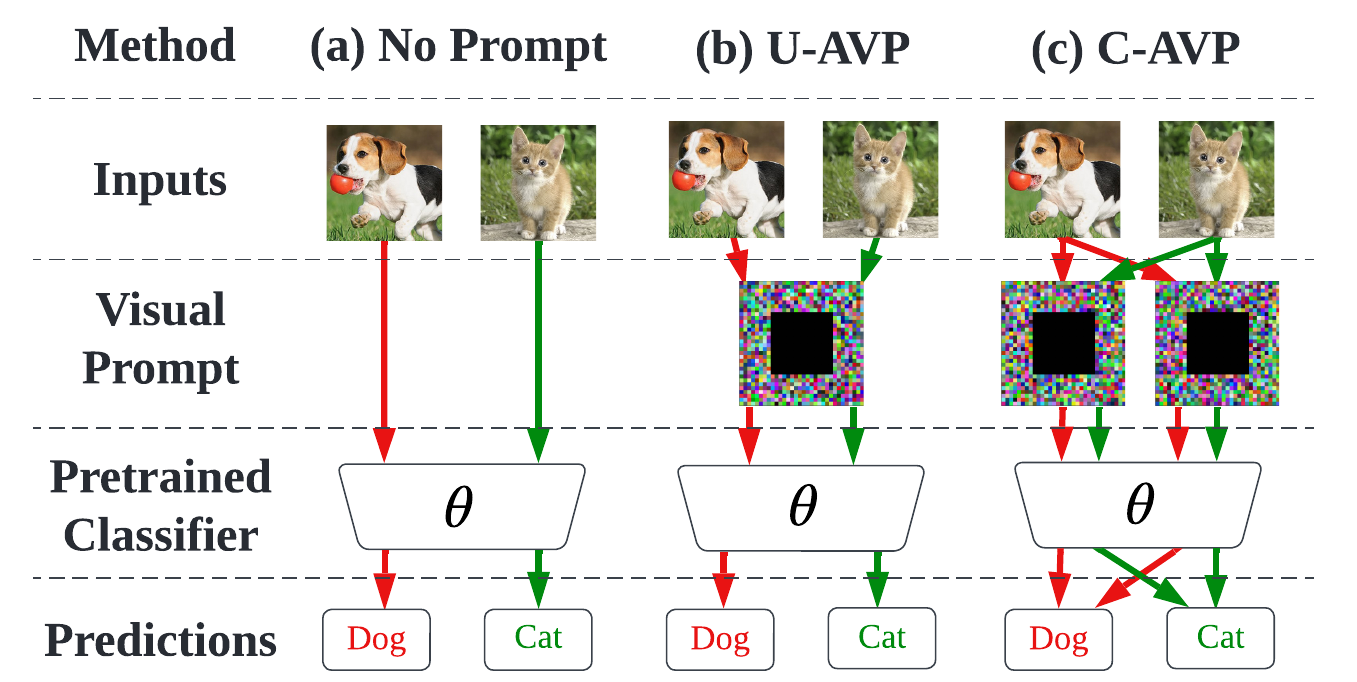}
    \caption{\footnotesize{Overview of C-AVP over two classes  (red and green) vs. \ref{eq: U_AVP}  and the prompt-free learning pipeline.  
    }}
    \label{fig: alg_illustration}
    \vspace*{-6mm}
\end{figure}

\noindent \textbf{Joint prompts optimization for C-AVP.}
The failure of  \ref{eq: C_AVP_vanilla}   inspires us to rethink the value of class-wise separability. As illustrated in challenges \textbf{(C1)}-\textbf{(C2)}, the compatibility with the test-time prompt selection rule  and  the interrelationship between class-wise visual prompts should be taken into account.  To this end, we develop a series of new AVP principles below. Fig.\,\ref{fig: alg_illustration} provides a schematic overview of {C-AVP} and its comparison with \ref{eq: U_AVP}  and the predictor without VP.

First, to bake the prompt selection rule  \eqref{eq: rule_prompt_sel} into C-AVP, we enforce the correct prompt selection, \textit{i.e.}, under the condition that 
$
f_y (\mathbf x+ \boldsymbol \delta^{(y)}; \btheta) >  \max_{k: k \neq y} f_k (\mathbf x+ \boldsymbol \delta^{(k)}; \btheta) 
$ for $(\mathbf x, y) \in \mathcal D^{(y)}$. The above can be cast as a CW-type loss 
\cite{carlini2017towards}:

{\vspace*{-3.5mm}
\small 
\begin{equation}
    \begin{aligned}
        & \ell_{\mathrm{C-AVP},1} (\{\boldsymbol \delta^{(i)} \} ; \mathcal D_\mathrm{tr}, \boldsymbol \theta) = \\  &\mathbb E_{(\mathbf x, y) \in \mathcal D_\mathrm{tr}}   \max \{ \max_{k \neq y}  f_k (\mathbf x+ \boldsymbol \delta^{(k)}; \btheta)  - f_y (\mathbf x+ \boldsymbol \delta^{(y)}; \btheta) , - \tau  \} ,
    \end{aligned}
    \label{eq: C-AVP_loss1}
\end{equation}
}%
where $\tau>0$ is a confidence threshold. The rationale behind \eqref{eq: C-AVP_loss1} is that given a data sample $(\mathbf x, y)$, the 
  minimum value of $ \ell_{\mathrm{C-AVP},1} $  is achieved at $-\tau$, indicating the desired condition  with the confidence level $\tau$. Compared with  \eqref{eq: C_AVP_vanilla}, another key characteristic of  $\ell_{\mathrm{C-AVP},1}$ is  its non-splitting over class-wise   prompts $\{\boldsymbol \delta^{(i)} \}$, which benefits the joint optimization of these prompts.   

Second, to mitigate the backdoor effect of mis-matched prompts, we propose additional two losses, noted by $\ell_{\mathrm{C-AVP},2}$ and $\ell_{\mathrm{C-AVP},3}$,  to penalize the data-prompt mismatches. 
Specifically, $\ell_{\mathrm{C-AVP},2}$  penalizes the  backdoor-alike targeted prediction accuracy of a class-wise visual prompt when applied to mismatched training data.
For the prompt $\boldsymbol \delta^{(i)}$, this leads to 

{\vspace*{-4.5mm}
\small 
\begin{equation}
    \begin{aligned}
        & \ell_{\mathrm{C-AVP},2} (\{\boldsymbol \delta^{(i)} \} ; \mathcal D_\mathrm{tr}, \boldsymbol \theta) = \\ &\frac{1}{N} \sum_{i=1}^N  \mathbb E_{(\mathbf x, y) \in \mathcal{D}_{\mathrm{tr}}^{(-i)}} \max \{  f_i (\mathbf x+ \boldsymbol \delta^{(i)}; \btheta) - f_y(\mathbf x + \boldsymbol \delta^{(i)}; \btheta) , - \tau  \} ,
    \end{aligned}
    \label{eq: C-AVP_loss2}
\end{equation}
}%
where $\mathcal{D}_{\mathrm{tr}}^{(-i)}$  denotes the training data set   by excluding $\mathcal{D}_{\mathrm{tr}}^{(i)}$.
The class $i$-associated prompt $\boldsymbol \delta^{(i)}$ should \textit{not} behave as a backdoor trigger to non-$i$ classes' data. Likewise, if the prompt is applied to the correct data class, then the prediction confidence should surpass that of a  mis-matched case. This leads to 

{\vspace*{-4.5mm}
\small 
\begin{equation}
    \begin{aligned}
        & \ell_{\mathrm{C-AVP},3} (\{\boldsymbol \delta^{(i)} \} ; \mathcal D_\mathrm{tr}, \boldsymbol \theta) = \\ &\mathbb E_{(\mathbf x, y) \in \mathcal D_\mathrm{tr}}  \max \{ \max_{k \neq y}  f_y (\mathbf x+ \boldsymbol \delta^{(k)}; \btheta)  - f_y (\mathbf x+ \boldsymbol \delta^{(y)}; \btheta) , - \tau  \}.
    \end{aligned}
    \label{eq: C-AVP_loss3}
\end{equation}
}%

Let $\ell_{\mathrm{C-AVP},0} (\{ \boldsymbol \delta^{(i)}\}; \mathcal{D}_\mathrm{tr}, \btheta)$ denote  the objective function of  \eqref{eq: C_AVP_vanilla}. Integrated with $\ell_{\mathrm{C-AVP}, q} (\{ \boldsymbol \delta^{(i)}\}; \mathcal{D}_\mathrm{tr}, \btheta)$ for $q \in \{1,2,3 \}$, the desired class-wise AVP design is cast as

{\vspace*{-4.5mm}
\small 
\begin{align}
    \begin{array}{ll}
        \displaystyle \minimize_{ \{ \boldsymbol \delta^{(i)} \in \mathcal C\}_{i\in [N]}} &  \ell_{\mathrm{C-AVP},0} (\{ \boldsymbol \delta^{(i)}\}; \mathcal{D}_\mathrm{tr}, \btheta) + \\ & \gamma \sum_{q=1}^3 \ell_{\mathrm{C-AVP},q} (\{ \boldsymbol \delta^{(i)}\}; \mathcal{D}_\mathrm{tr}, \btheta),
    \end{array}
    \label{eq: C_AVP_CW}
    \tag{C-AVP}
\end{align}}%
where $\gamma > 0$ is a parameter for class-wise prompting penalties. 

\vspace*{-4mm}
\section{Experiments}
\begin{figure*}[htb]
    \centering
    \includegraphics[width=.8\textwidth]{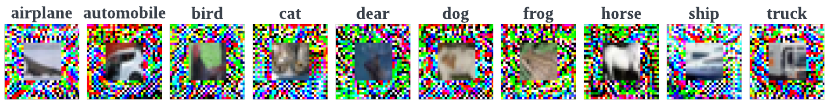}
    \vspace*{-4mm}
    \caption{\footnotesize{C-AVP visualization. One image is chosen from each CIFAR-10 class with the corresponding C-AVP.}}
    \label{fig: C-AVP visualization}
    \vspace*{-4mm}
\end{figure*}

\vspace*{-2mm}
\noindent \textbf{Experiment setup.}
We conduct experiments on CIFAR-10 with a pretrained ResNet18 of testing accuracy of 94.92\% on  standard test dataset. We use PGD-10
(\textit{i.e.}, PGD attack with 10 steps \cite{madry2017towards}) to
generate adversarial examples with $\epsilon=8/255$ during visual prompts  training, and with a cosine learning rate scheduler   starting at $0.1$. Throughout experiments, we choose $\lambda=1$  in \eqref{eq: U_AVP}, and $\tau=0.1$ and $\gamma=3$ in \eqref{eq: C_AVP_CW}. The width of visual prompt  is set to 8 (see Fig.\,\ref{fig: C-AVP visualization} for the visualization).

\begin{table}
\centering
\caption{\footnotesize{VP performance comparison in terms of standard (std) accuracy (acc)  and robust accuracy against PGD attacks with  $\epsilon=8/255$  and multiple PGD  steps on (CIFAR-10, ResNet18).} }
\label{tab: ra_sa_stat}
\resizebox{0.35\textwidth}{!}{%
\begin{tabular}{c|c|cccc}
\toprule[1pt]
\midrule
Evaluation  & Std  &\multicolumn{4}{c}{Robust acc vs PGD w/ step \#} \\ 
metrics (\%) & acc & 10 & 20 & 50 & 100 \\ 
\midrule

Pre-trained 
& \textbf{94.92}
& 0
& 0
& 0
& 0
\\

Vanilla VP 
& 94.48
& 0
& 0
& 0
& 0
\\

\ref{eq: U_AVP}
& 27.75
& 16.9
& 16.81
& 16.81
& 16.7
\\

\ref{eq: C_AVP_vanilla}
& 19.69
& 13.91
& 13.63
& 13.6
& 13.58
\\

\rowcolor[gray]{.8}
\ref{eq: C_AVP_CW} (ours)
& 57.57
& \textbf{34.75}
& \textbf{34.62}
& \textbf{34.51}
& \textbf{33.63}

\\
\midrule
\bottomrule[1pt]
\end{tabular}%
}
\vspace*{-3mm}
\end{table}

\noindent \textbf{\ref{eq: C_AVP_CW} outperforms conventional VP.}
Tab.\,\ref{tab: ra_sa_stat} demonstrates the effectiveness of   proposed \ref{eq: C_AVP_CW} approach vs. \ref{eq: U_AVP} (the direct extension of VP to adversarial defense) and the \ref{eq: C_AVP_vanilla} method in the task of robustify a normally-trained ResNet18 on CIFAR-10. For comparison, we also report the standard accuracy of the pre-trained model and the vanilla VP solution given by \eqref{eq: visual_prompt}.
As we can see, \ref{eq: C_AVP_CW}
outperforms U-AVP and C-AVP-v0 in both standard accuracy and robust accuracy.
We also observe that compared to the pretrained model and the vanilla VP, the robustness-induced VP variants bring in an evident standard accuracy drop as the cost of robustness.

\noindent \textbf{Prompting regularization effect in \eqref{eq: C_AVP_CW}.} Tab.\,\ref{tab: AVP-ablation} shows different settings of prompting regularizations used in C-AVP, where `S$i$' represents a  certain loss configuration.  As we can see, the use of $ \ell_{\mathrm{C-AVP},2}$ contributes most to the performance of learned visual prompts (see S$3$). This is not surprising, since we design $ \ell_{\mathrm{C-AVP},2}$ for mitigating the backdoor effect of class-wise prompts, which is the main source of prompting selection error. We also note that $ \ell_{\mathrm{C-AVP},1}$ is the second most important regularization. This is because such a regularization is accompanied with the prompt selection rule \eqref{eq: rule_prompt_sel}. Tab.\,\ref{tab: AVP-ablation} also indicates that the combination of $ \ell_{\mathrm{C-AVP},1}$ and $ \ell_{\mathrm{C-AVP},2}$ is a possible computationally lighter alternative  to \eqref{eq: C_AVP_CW}.

\begin{figure}[htb]
\centering
\includegraphics[width=\columnwidth]{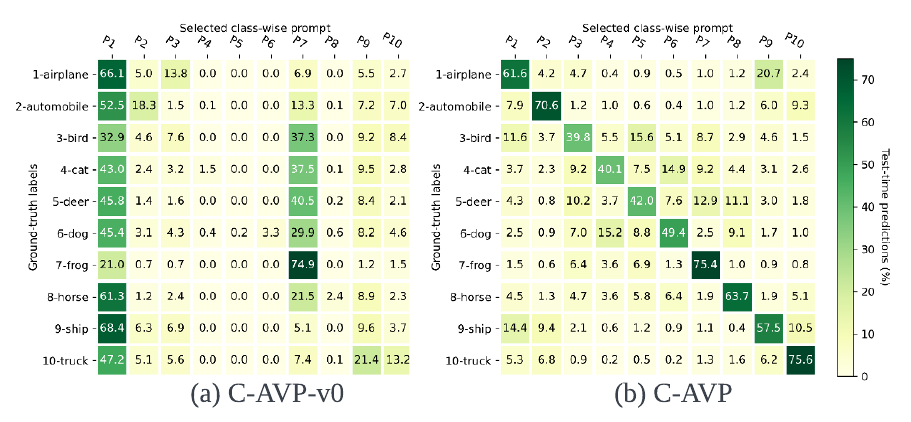}
\caption{\footnotesize{The predictions of \ref{eq: C_AVP_vanilla} vs. \ref{eq: C_AVP_CW} on (CIFAR10, ResNet18). }}
\label{fig: test_predictions}
\vspace*{-4mm}
\end{figure}

\noindent \textbf{Class-wise prediction error analysis.}
Fig.\,\ref{fig: test_predictions} shows a comparison of the classification confusion matrix. Each row corresponds to testing samples from one class, and each column corresponds to the prompt (`P')  selection across 10 image classes.  As we can see, our proposal outperforms  \ref{eq: C_AVP_vanilla} since the former's higher main diagonal entries indicate less prompt selection error than the latter.

\noindent \textbf{Comparisons with other test-time defenses.} 
In Tab.\,\ref{tab:ressota}, we compare our proposed \ref{eq: C_AVP_CW} with three test-time defense methods selected from Croce \textit{et. al.}\cite{croce2022evaluating}. Note that all methods are applied to robustifying a fixed, standardly pre-trained ResNet18. Following Croce \textit{et. al.}\cite{croce2022evaluating}, we divide the considered defenses into different categories, relying on their defense principles (\textit{i.e.}, IP or MA) and needed test-time operations  (\textit{i.e.}, IA, AN, and R). As we can see, our method \ref{eq: C_AVP_CW} falls into the IP category but requires no involved test-time operations. This leads to the least inference overhead. Although there exists a performance gap with the test-time defense baselines, we hope that our work could pave a way to study the pros and cons of visual prompting in adversarial robustness.  

\vspace*{-4mm}
\begin{table}
\centering
\caption{\footnotesize{Sensitivity analysis of prompting regularization in \ref{eq: C_AVP_CW} on (CIFAR-10, ResNet18).} }
\vspace*{-2mm}
\label{tab: AVP-ablation}
\resizebox{\columnwidth}{!}{%
\begin{tabular}{c|ccc|c|c}
\toprule[1pt]
\midrule
Setting&$\ell_{\mathrm{C-AVP},1}$&$ \ell_{\mathrm{C-AVP},2}$ & $\ell_{\mathrm{C-AVP},3}$ & Std Acc (\%) & PGD-10 Acc (\%)  \\ 
\midrule

S$1$ &
 \redcross
&  \redcross
&  \redcross
& 19.69

& 13.91

\\
S$2$ &
 \greencheck
&  \redcross
&  \redcross
& 22.72
& 13.01
\\

S$3$ &
 \redcross
& \greencheck
&  \redcross
& 40.01
& 25.40

\\

S$4$ &
 \redcross
&  \redcross
& \greencheck
& 17.44

& 11.78

\\

S$5$ &
\greencheck
& \greencheck
& \redcross
& 57.03

& 32.39

\\

S$6$ &
 \greencheck
&  \redcross
& \greencheck
& 26.02
& 15.80

\\

\rowcolor[gray]{.8}
S$7$ &
\greencheck
& \greencheck
& \greencheck
& \textbf{57.57}
& \textbf{34.75}
\\


\midrule
\bottomrule[1pt]
\end{tabular}%
}
\vspace*{-3mm}
\end{table}

\begin{table}[htb]
\centering
\caption{\footnotesize{Comparison of \ref{eq: C_AVP_CW} with other SOTA test-time defenses. Per the benchmark in Croce \textit{et. al.}\cite{croce2022evaluating},  the involved test-time operations in these defenses include: IP (input purification), MA (model adaption),  IA (iterative algorithm),  AN (auxiliary network),  and  R (randomness). And  inference time (IT),  standard accuracy (SA), and robust accuracy (RA) against PGD-10  are used as performance metrics.} }
\label{tab:ressota}
\vspace*{-2mm}
\resizebox{0.4\textwidth}{!}{%
\begin{tabular}{c|ccccc|c|c|c}
\toprule[1pt]
\midrule
Method&IP&MA & IA & AN & R & IT & SA (\%) & RA (\%)  \\ 
\midrule

\cite{shi2021online} &
 \greencheck
&  \redcross
&  \greencheck
&  \redcross
&  \redcross
& 518 $\times$
& 85.9\%
& 0.4\%
\\

\cite{yoon2021adversarial} &
\greencheck
&  \redcross
& \greencheck
& \greencheck
& \greencheck
& 176 $\times$
& 91.1\%
& 40.3\%
\\

\cite{chen2021towards}  &
\redcross 
&  \greencheck
&  \greencheck
&  \greencheck
&  \redcross
& 59 $\times$
& 56.1\%
& 50.6\%
\\

\rowcolor[gray]{.8}
\ref{eq: C_AVP_CW} &
 \greencheck
&  \redcross
&  \redcross
&  \redcross
&  \redcross
& \textbf{1.4 $\times$}
& 57.6\%
& 34.8\%
\\

\midrule
\bottomrule[1pt]
\end{tabular}%
}
\vspace*{-4mm}
\end{table}

\vspace*{-4mm}
\section{Conclusion}
\vspace{-3mm}
In this work, we develop a novel VP method, \textit{i.e.},  \ref{eq: C_AVP_CW}, to improve adversarial robustness of a fixed model at test time. Compared to existing VP methods, this is the first work to peer into how VP could be in adversarial defense. We show the direct integration of VP into robust learning is \textit{not} an effective adversarial defense at test time for a fixed model. To address this problem, we propose \ref{eq: C_AVP_CW} to create ensemble visual prompts and  jointly optimize their interrelations for robustness enhancement. We empirically show that our proposal significantly reduces the inference overhead compared to classical adversarial defenses which typically call for computationally-intensive test-time defense operations.

{\small 
\bibliographystyle{IEEEbib}
\bibliography{refs}

\begin{thebibliography}{10}

\bibitem{madry2017towards}
Aleksander Madry et~al.,
\newblock ``Towards deep learning models resistant to adversarial attacks,''
\newblock {\em arXiv:1706.06083}, 2017.

\bibitem{zhang2019theoretically}
Hongyang Zhang, Yaodong Yu, et~al.,
\newblock ``Theoretically principled trade-off between robustness and
  accuracy,''
\newblock {\em ICML}, 2019.

\bibitem{shafahi2019adversarial}
Ali Shafahi, Mahyar Najibi, et~al.,
\newblock ``Adversarial training for free!,''
\newblock in {\em NeurIPS}, 2019.

\bibitem{zhang2019you}
Dinghuai Zhang, Tianyuan Zhang, et~al.,
\newblock ``You only propagate once: Accelerating adversarial training via
  maximal principle,''
\newblock {\em arXiv:1905.00877}, 2019.

\bibitem{carmon2019unlabeled}
Yair Carmon, Aditi Raghunathan, et~al.,
\newblock ``Unlabeled data improves adversarial robustness,''
\newblock {\em NeurIPS}, 2019.

\bibitem{wong2017provable}
Eric Wong and J~Zico Kolter,
\newblock ``Provable defenses against adversarial examples via the convex outer
  adversarial polytope,''
\newblock {\em arXiv:1711.00851}, 2017.

\bibitem{raghunathan2018certified}
Aditi Raghunathan, Jacob Steinhardt, et~al.,
\newblock ``Certified defenses against adversarial examples,''
\newblock {\em arXiv:1801.09344}, 2018.

\bibitem{xie2019feature}
Cihang Xie, Yuxin Wu, et~al.,
\newblock ``Feature denoising for improving adversarial robustness,''
\newblock in {\em CVPR}, 2019.

\bibitem{chen2020adversarial}
Tianlong Chen, Sijia Liu, et~al.,
\newblock ``Adversarial robustness: From self-supervised pre-training to
  fine-tuning,''
\newblock in {\em CVPR}, 2020.

\bibitem{fan2021does}
Lijie Fan, Sijia Liu, et~al.,
\newblock ``When does contrastive learning preserve adversarial robustness from
  pretraining to finetuning?,''
\newblock {\em NeurIPS}, 2021.

\bibitem{jia2022clawsat}
Jinghan Jia et~al.,
\newblock ``Clawsat: Towards both robust and accurate code models,''
\newblock {\em arXiv:2211.11711}, 2022.

\bibitem{athalye2018obfuscated}
Anish Athalye, Nicholas Carlini, et~al.,
\newblock ``Obfuscated gradients give a false sense of security: Circumventing
  defenses to adversarial examples,''
\newblock {\em arXiv:1802.00420}, 2018.

\bibitem{croce2020reliable}
Francesco Croce and Matthias Hein,
\newblock ``Reliable evaluation of adversarial robustness with an ensemble of
  diverse parameter-free attacks,''
\newblock in {\em ICML}. PMLR, 2020.

\bibitem{croce2022evaluating}
Francesco Croce et~al.,
\newblock ``Evaluating the adversarial robustness of adaptive test-time
  defenses,''
\newblock {\em arXiv:2202.13711}, 2022.

\bibitem{yoon2021adversarial}
Jongmin Yoon, Sung~Ju Hwang, et~al.,
\newblock ``Adversarial purification with score-based generative models,''
\newblock in {\em ICML}. PMLR, 2021.

\bibitem{mao2021adversarial}
Chengzhi Mao, Mia Chiquier, et~al.,
\newblock ``Adversarial attacks are reversible with natural supervision,''
\newblock in {\em ICCV}, 2021.

\bibitem{alfarra2022combating}
Motasem Alfarra, Juan~C P{\'e}rez, et~al.,
\newblock ``Combating adversaries with anti-adversaries,''
\newblock in {\em AAAI}, 2022.

\bibitem{salman2020denoised}
Hadi Salman, Mingjie Sun, et~al.,
\newblock ``Denoised smoothing: A provable defense for pretrained
  classifiers,''
\newblock {\em NeurIPS}, 2020.

\bibitem{gong2022reverse}
Yifan Gong et~al.,
\newblock ``Reverse engineering of imperceptible adversarial image
  perturbations,''
\newblock {\em arXiv:2203.14145}, 2022.

\bibitem{kang2021stable}
Qiyu Kang et~al.,
\newblock ``Stable neural ode with lyapunov-stable equilibrium points for
  defending against adversarial attacks,''
\newblock {\em NeurIPS}, 2021.

\bibitem{bahng2022visual}
Hyojin Bahng et~al.,
\newblock ``Visual prompting: Modifying pixel space to adapt pre-trained
  models,''
\newblock {\em arXiv:2203.17274}, 2022.

\bibitem{chen2022model}
Pin-Yu Chen,
\newblock ``Model reprogramming: Resource-efficient cross-domain machine
  learning,''
\newblock {\em arXiv:2202.10629}, 2022.

\bibitem{elsayed2018adversarial}
Gamaleldin~F Elsayed, Ian Goodfellow, et~al.,
\newblock ``Adversarial reprogramming of neural networks,''
\newblock {\em arXiv:1806.11146}, 2018.

\bibitem{tsai2020transfer}
Yun-Yun Tsai et~al.,
\newblock ``Transfer learning without knowing: Reprogramming black-box machine
  learning models with scarce data and limited resources,''
\newblock {\em arXiv:2007.08714}, 2020.

\bibitem{zhang2022fairness}
Guanhua Zhang, Yihua Zhang, et~al.,
\newblock ``Fairness reprogramming,''
\newblock {\em arXiv:2209.10222}, 2022.

\bibitem{salman2021unadversarial}
Hadi Salman, Andrew Ilyas, et~al.,
\newblock ``Unadversarial examples: Designing objects for robust vision,''
\newblock {\em NeurIPS}, 2021.

\bibitem{brown2020language}
Tom Brown, Benjamin Mann, et~al.,
\newblock ``Language models are few-shot learners,''
\newblock {\em NeurIPS}, 2020.

\bibitem{li2021prefix}
Xiang~Lisa Li and Percy Liang,
\newblock ``Prefix-tuning: Optimizing continuous prompts for generation,''
\newblock {\em arXiv:2101.00190}, 2021.

\bibitem{radford2021learning}
Alec Radford et~al.,
\newblock ``Learning transferable visual models from natural language
  supervision,''
\newblock in {\em ICML}. PMLR, 2021.

\bibitem{zhang2023text}
Yimeng Zhang et~al.,
\newblock ``Text-visual prompting for efficient 2d temporal video grounding,''
\newblock {\em arXiv:2303.04995}, 2023.

\bibitem{neekhara2022cross}
Paarth Neekhara, Shehzeen Hussain, et~al.,
\newblock ``Cross-modal adversarial reprogramming,''
\newblock in {\em WACV}, 2022.

\bibitem{yang2021voice2series}
Chao-Han~Huck Yang, Yun-Yun Tsai, et~al.,
\newblock ``Voice2series: Reprogramming acoustic models for time series
  classification,''
\newblock in {\em ICML}. PMLR, 2021.

\bibitem{zheng2021adversarial}
Yang Zheng, Xiaoyi Feng, et~al.,
\newblock ``Why adversarial reprogramming works, when it fails, and how to tell
  the difference,''
\newblock {\em arXiv:2108.11673}, 2021.

\bibitem{grosse2017statistical}
Kathrin Grosse, Praveen Manoharan, et~al.,
\newblock ``On the (statistical) detection of adversarial examples,''
\newblock {\em arXiv:1702.06280}, 2017.

\bibitem{yang2019ml}
Puyudi Yang et~al.,
\newblock ``Ml-loo: Detecting adversarial examples with feature attribution,''
\newblock {\em arXiv:1906.03499}, 2019.

\bibitem{metzen2017detecting}
Jan~Hendrik Metzen, Tim Genewein, et~al.,
\newblock ``On detecting adversarial perturbations,''
\newblock {\em arXiv:1702.04267}, 2017.

\bibitem{meng2017magnet}
Dongyu Meng and Hao Chen,
\newblock ``Magnet: a two-pronged defense against adversarial examples,''
\newblock {\em arXiv:1705.09064}, 2017.

\bibitem{wojcik2020adversarial}
Bartosz W{\'o}jcik, Pawe{\l} Morawiecki, et~al.,
\newblock ``Adversarial examples detection and analysis with layer-wise
  autoencoders,''
\newblock {\em arXiv:2006.10013}, 2020.

\bibitem{boopathy2020proper}
Akhilan Boopathy et~al.,
\newblock ``Proper network interpretability helps adversarial robustness in
  classification,''
\newblock in {\em ICML}, 2020.

\bibitem{ye2019adversarial}
Shaokai Ye, Kaidi Xu, et~al.,
\newblock ``Adversarial robustness vs model compression, or both?,''
\newblock {\em arXiv e-prints}, 2019.

\bibitem{mohapatra2020rethinking}
Jeet Mohapatra, Ching-Yun Ko, et~al.,
\newblock ``Rethinking randomized smoothing for adversarial robustness,''
\newblock {\em arXiv:2003.01249}, 2020.

\bibitem{wang2021on}
Ren Wang, Kaidi Xu, et~al.,
\newblock ``On fast adversarial robustness adaptation in model-agnostic
  meta-learning,''
\newblock in {\em ICLR}, 2021.

\bibitem{shi2021online}
Changhao Shi, Chester Holtz, et~al.,
\newblock ``Online adversarial purification based on self-supervision,''
\newblock {\em arXiv:2101.09387}, 2021.

\bibitem{chen2021towards}
Zhuotong Chen, Qianxiao Li, et~al.,
\newblock ``Towards robust neural networks via close-loop control,''
\newblock {\em arXiv:2102.01862}, 2021.

\bibitem{zhang2022robustify}
Yimeng Zhang, Yuguang Yao, et~al.,
\newblock ``How to robustify black-box ml models? a zeroth-order optimization
  perspective,''
\newblock {\em arXiv:2203.14195}, 2022.

\bibitem{gu2017badnets}
Tianyu Gu, Brendan Dolan-Gavitt, et~al.,
\newblock ``Badnets: Identifying vulnerabilities in the machine learning model
  supply chain,''
\newblock {\em arXiv:1708.06733}, 2017.

\bibitem{carlini2017towards}
Nicholas Carlini et~al.,
\newblock ``Towards evaluating the robustness of neural networks,''
\newblock in {\em IEEE Symposium on S\&P}, 2017.

\end{thebibliography}
}

\end{document}